\def\year{2019}\relax
\documentclass[letterpaper]{article} 
\usepackage{aaai19}  
\usepackage{times}  
\usepackage{helvet} 
\usepackage{courier}  
\usepackage[hyphens]{url}  
\usepackage{graphicx} 
\usepackage{graphicx}  
\usepackage[table]{xcolor}

\title{``Superstition'' in the Network:\\ 
Deep Reinforcement Learning Plays Deceptive Games}

\author{Philip Bontrager,\textsuperscript{\rm 1} 
Ahmed Khalifa,\textsuperscript{\rm 1} 
Damien Anderson,\textsuperscript{\rm 2} 
Matthew Stephenson,\textsuperscript{\rm 3} \\ 
\Large \textbf{
Christoph Salge, \textsuperscript{\rm 4}
Julian Togelius \textsuperscript{\rm 1}}\\
\textsuperscript{1}New York University,
\textsuperscript{2}University of Strathclyde,
\textsuperscript{3}Maastricht University,
\textsuperscript{4}University of Hertfordshire\\
  \{philipjb, ahmed.khalifa\}@nyu.edu,
  damien.anderson@strath.ac.uk,
  matthew.stephenson@maastrichtuniversity.nl,\\
  ChristophSalge@gmail.com,
  julian@togelius.com
}
\begin{document}

\maketitle

\begin{abstract}
Deep reinforcement learning has learned to play many games well, but failed on others. To better characterize the modes and reasons of failure of deep reinforcement learners, we test the widely used Asynchronous Actor-Critic (A2C) algorithm on four deceptive games, which are specially designed to provide challenges to game-playing agents. These games are implemented in the General Video Game AI framework, which allows us to compare the behavior of reinforcement learning-based agents with planning agents based on tree search. We find that several of these games reliably deceive deep reinforcement learners, and that the resulting behavior highlights the shortcomings of the learning algorithm. The particular ways in which agents fail differ from how planning-based agents fail, further illuminating the character of these algorithms. We propose an initial typology of deceptions which could help us better understand pitfalls and failure modes of (deep) reinforcement learning.
\end{abstract}

\section{Introduction}

In reinforcement learning (RL) \cite{sutton1998reinforcement} an agent is tasked with learning a policy that maximizes expected reward based only on its interactions with the environment. 
In general, there is no guarantee that any such procedure will lead to an optimal policy; while convergence proofs exist, they only apply to a tiny and rather uninteresting class of environments. Reinforcement learning still performs well for a wide range of scenarios not covered by those convergence proofs. However, while recent successes in game-playing with deep reinforcement learning~\cite{justesen2017deep} have led to a high degree of confidence in the deep RL approach, there are still scenarios or games where deep RL fails.
Some oft-mentioned reasons why RL algorithms fail are partial observability and long time spans between actions and rewards.
But are there other causes?   

In this paper, we want to address these questions by looking at games that are designed to be deliberately deceptive. Deceptive games are defined as those where the reward structure is designed to lead away from an optimal policy. For example, games where learning to take the action which produces early rewards curtails further exploration. Deception does not include outright lying (or presenting false information). More generally speaking, deception is the exploitation of cognitive biases. Better and faster AIs have to make some assumptions to improve their performance or generalize over their observation (as per the no free lunch theorem, an algorithm needs to be tailored to a class of problems in order to improve performance on those problems~\cite{wolpert1997no}). These assumptions in turn make them susceptible to deceptions that subvert these very assumptions. For example, evolutionary optimization approaches assume locality, i.e., that solutions that are close in genome space have a similar fitness - but if very bad solutions surround a very good solution, then an evolutionary algorithm would be less likely to find it than random search. 

While we are specifically looking at digital games here, the ideas we discuss are related to the question of optimization and decision making in a broader context. Many real-world problems involve some form of deception; for example, while eating sugar brings momentary satisfaction, a long-term policy of eating as much sugar as possible is not optimal in terms of health outcomes. 

In a recent paper, a handful of \emph{deceptive games} were proposed, and the performance of a number of planning algorithms were tested on them~\cite{anderson2018deceptive}. It was shown that many otherwise competent game-playing agents succumbed to these deceptions and that different types of deceptions affected different kinds of planning algorithms; for example, agents that build up a model of the effects of in-game objects are vulnerable to deceptions based on changing those effects. In this paper, we want to see how well deep reinforcement learning performs on these games. This approach aims to gain a better understanding of the vulnerabilities of deep reinforcement learning. 

\section{Background}

Reinforcement learning algorithms learn through interacting with an environment and receiving rewards~\cite{sutton1998reinforcement}. There are different types of algorithms that fit this bill. A core distinction between the types are between ontogenetic algorithms, that learn within episodes from the reward that they encounter, and phylogenetic algorithms, that learn between episodes based on the aggregate reward at the end of each episode~\cite{togelius2009ontogenetic}. 

For some time, reinforcement learning had few clear successes. However, in the last five years, the combination of ontogenetic RL algorithms with deep neural networks have seen significant successes, in particular in playing video games~\cite{justesen2017deep} such as simple 2D arcade games~\cite{mnih2015human} to more advanced games like Dota 2 and Starcraft \cite{OpenAI_dota,alphastarblog}. This combination, generally referred to as deep reinforcement learning, is the focus of much research.

The deceptive games presented in this paper were developed for the GVGAI (General Video Game Artificial Intelligence~\cite{perez2016general}) framework. The GVGAI framework itself is based on VGDL (Video Game Description Language~\cite{ebner2013towards,Schaul2013}) which is a language that was developed to express a range of arcade games, like Sokoban and Space Invaders. VGDL was developed to encourage research into more general video game playing~\cite{levine2013general} by providing a language and an interface to a range of arcade games. 
Currently the GVGAI corpus has over 150 games. The deceptive games discussed in this paper are fully compatible with the framework. 

\section{Methods}

To empirically test the effectiveness of the deception in every game, we train a reinforcement learning algorithm and run six planning algorithms on each game. The benefit of working in GVGAI is that we are able to evaluate the same game implementations with algorithms that require an available forward model and with learning agents. GVGAI has a Java interface for planning agents as well as an OpenAI Gym interface for learning agents \cite{perez2016general,torrado2018deep,brockman2016openai}. 

All algorithms were evaluated on each game 150 times. The agent's scores are evaluated along with play through videos. The qualitative analyses of the videos provide key insights into the causes behind certain scores and into what an agent is actually learning. The quantitative and qualitative results are then used for the final analysis.

\subsection{Reinforcement Learning}

To test if these games are capable of deceiving an agent trained via reinforcement learning, we use Advantage Actor-Critic (A2C) to learn to play the games \cite{mnih2016asynchronous}. A2C is a good benchmark algorithm and has been shown to be capable of playing GVGAI games with some success \cite{torrado2018deep,justesen2018procedural}. A2C is a model-free,extrinsically driven algorithm that allows for examining the effects of different reward patterns. A2C is also relevant due to the popularity of model-free agents.

Due to the arcade nature of GVGAI games, we train on pixels with the same setup developed for the Atari Learning Environment framework \cite{bellemare13arcade}. The atari configuration has been shown to work well for GVGAI and allows a consistent baseline with which to compare all the games \cite{torrado2018deep}. Instead of tuning the algorithms for the games, we designed the games for the algorithms. We use the OpenAI Baselines implementation of A2C \cite{baselines}. The neural network architecture is the same as the original designed by Mnih et al. \cite{mnih2016asynchronous}. The hyper-parameters are the default from the original paper as implemented by OpenAI: step size of 5, no frame skipping, constant learning rate of 0.007, RMS, and we used 12 workers.

For each environment, we trained five different A2C agents to play, each starting from random seeds. In initial testing, we tried training for twenty million frames, and we found that the agents converged very quickly, normally within two million frames of training. We therefore standardized the experiments to all train for five million frames. One stochastic environment, WaferThinMints, did not converge and might have benefited from more training time.

\subsection{Planning Agents}

For comparison with previous work and better insight into the universality of the deceptive problems posed here, we compare our results to planning algorithms. What we mean by planning agents are algorithms that utilize a forward model to search for an ideal game state. In the GVGAI planning track, each algorithm is provided with the current state and a forward model and it has to return the next action in a small time frame (40 milliseconds). This time frame doesn't give the algorithm enough time to find the best action. This limitation forces traditional planning algorithms to be somewhat greedy which, for most of these games, is a trap. 

In this paper, we are using six different planning algorithms. Three of them (aStar, greedySearch, and sampleMCTS) are directly from the GVGAI framework, while the rest (NovelTS, Return42, and YBCriber) are collected from the previous GVGAI competitions. Two of these algorithms, Return42, and YBCriber, are hybrid algorithms. They use one approach for deterministic games, such as A* or Iterative Width, and a different one for stochastic games, such as random walk or MCTS. Both algorithms use hand designed heuristics to judge game states. These hybrid algorithms also use online learning to bypass the small time per frame. The online learning agents try to understand the game rules, from the forward model during each time step, and then use that knowledge to improve the search algorithm.

\section{Deceptive Games}
In our previous work, a suite of deceptive games was created in order to take a look at the effects that these deceptive mechanics would have on agents \cite{anderson2018deceptive}. These deceptive games were designed in order to deceive different types of agents in different ways. 

From a game design perspective, the category of deceptive games partially overlaps with ``abusive games'', as defined by Wilson and Sicart~\cite{wilson2010now}. In particular, the abuse modalities of ``unfair design'' can be said to apply to some of the games we describe below. Wilson and Sicart note that these modalities are present in many commercial games, even successful and beloved games, especially those from the 8-bit era.

This section describes some of these games in detail, and defines optimal play for an agent playing each game. We focus on four key categories of deception that these games exploit. We believe these categories represent general problems that learning agents face and these simple games allow us to shine a spotlight on weaknesses that model-free, deep reinforcement learning agents still face. For a more comprehensive list of types of deceptions and deceptive games see Deceptive Games \cite{anderson2018deceptive}.

The following four different categories of deception will be discussed further in the discussion section: Lack of Hierarchical Understanding, Subverted Generalization, Delayed Gratification, and Delayed Reward.

\subsection{DeceptiCoins (DC)}

\begin{figure}[t]
\centering
\includegraphics[width=1\linewidth]{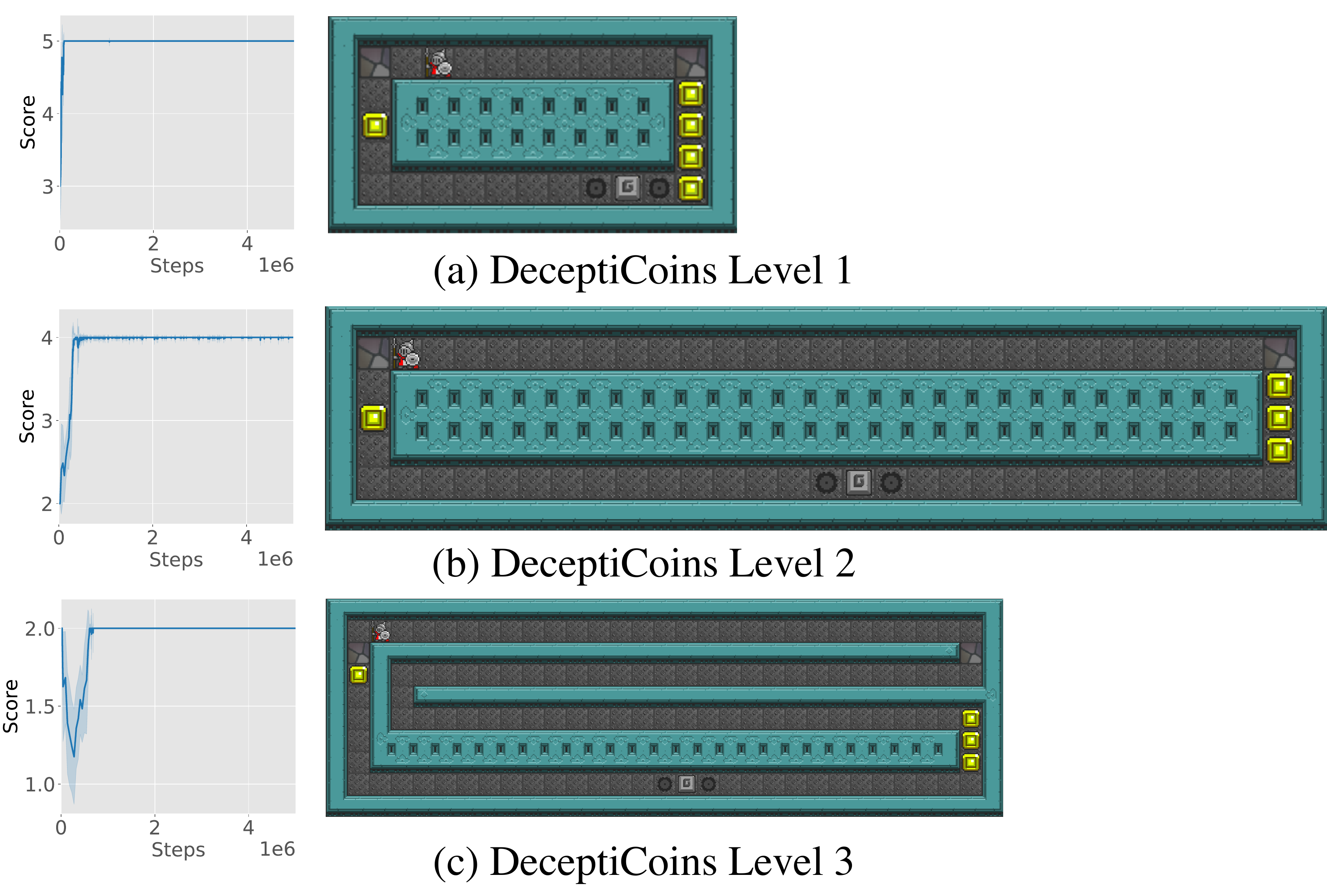}
\caption{DeceptiCoins Levels}
\label{fig:dc}
\end{figure}

\subsubsection{Game} DeceptiCoins, Figure \ref{fig:dc}, offers an agent two paths which both lead to the win condition. The first path presents immediate points to the agent, in the form of gold coins. The second path contains more gold coins, but they are further away and may not be immediately visible to a short-sighted agent. Once the agent selects a path, they become trapped within their chosen path and can only continue to the bottom exit. The levels used here are increasingly larger versions of the same challenge, but remain relatively small overall.



The optimal strategy for DeceptiCoins is to select the path with the highest overall number of points. For the levels shown in Figure 1, this is achieved by taking the right side path, as it leads to the highest total score (i.e., more gold coins can be collected before completing the level).


\subsubsection{Goal}
The game offers a simple form of deception that targets the exploration versus exploitation problem that learning algorithms face. The only way for the learning agent to discover the higher reward is for it to forgo the natural reward it discovers early on completely. By designing different sized levels, we can see how quickly the exploration space becomes too large. 
At the same time, an agent that correctly learns, on the short route, about coins and navigation could then see that going right is superior.

\subsubsection{Results} 
The first two levels of DeceptiCoins are very small, and the agent fairly quickly learns the optimal strategy. However, In level two the agent took several times longer to discover the optimal strategy, as expected from an agent that can only look at the rewards of individual moves. Level 3 proves to be too hard, and the agent converges on the suboptimal strategy. By comparison, a randomly initialized agent is very likely to select the easy path, since it starts next to it, before being forced to move toward the exit.

The training curve for level 3 shows a significant drop in performance at the beginning of training. The video footage suggests that the agent learns the concept of the gold coins and is attempting to collect them all, but fails to understand that once it takes the easy coin it will become trapped in the left path. The agent will also move back and forth between the paths at the beginning of the game, trying to decide.

\subsection{WaferThinMints (Mints)}

\begin{figure}[t!]
    \centering
    \includegraphics[width=.7\linewidth]{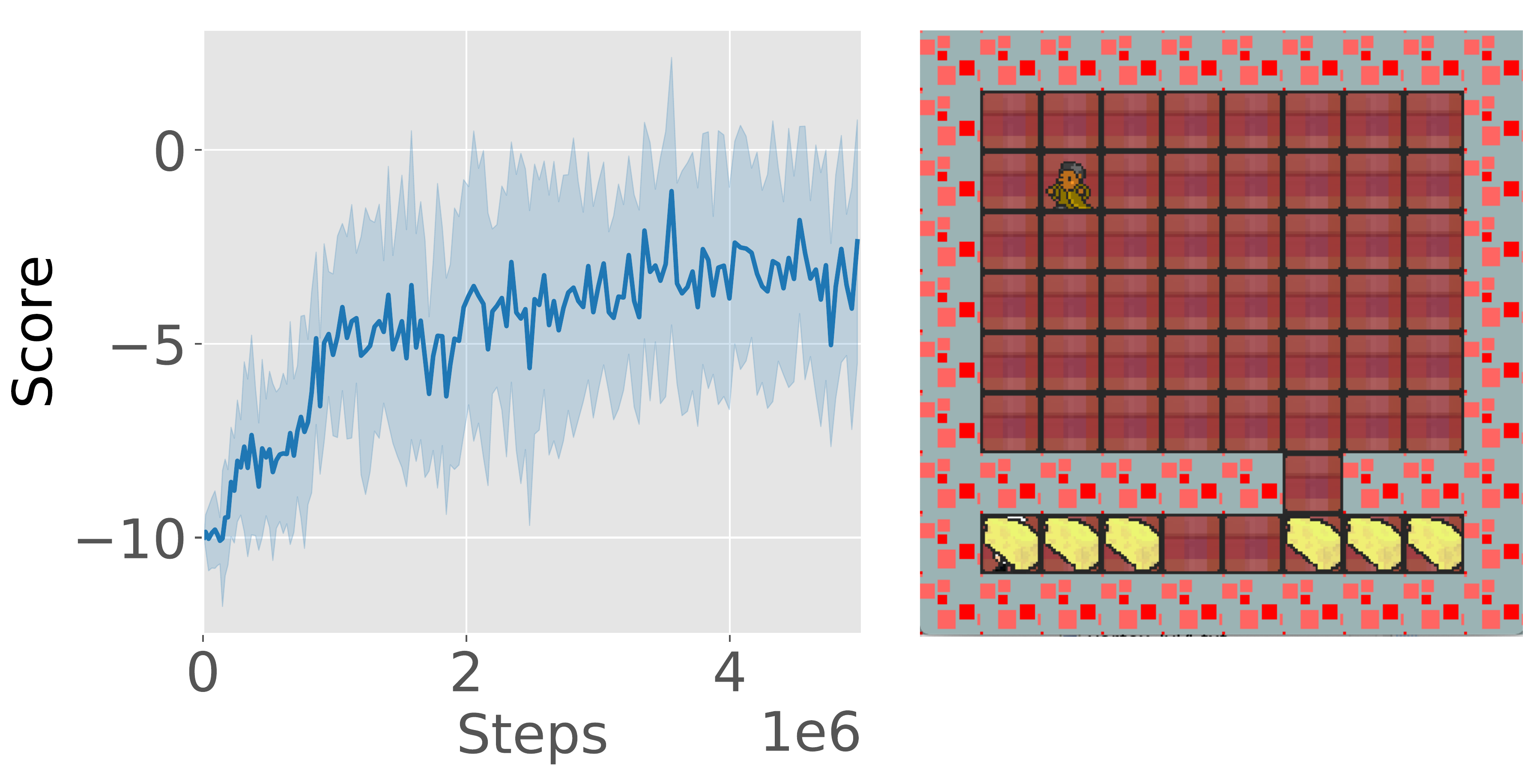}
    \caption{The first level of WaferThinMints}
    \label{fig:wfmLevel1}
\end{figure}

\subsubsection{Game} WaferThinMints is inspired by a scene in Monty Python's \emph{The Meaning of Life}. The game presents the agent with easily obtainable points, but if the agent collects too many it will lead to a loss condition. The idea of this game is to model a situation where a repeated action does not always lead to the same outcome or has a diminishing return over time. The levels for this game feature mints which each award a point when collected and also fill up a resource gauge on the agent. The level used is shown in figure \ref{fig:wfmLevel1}. If the avatar's resource gauge (green bar on avatar) is filled, defined in this case as nine mints, and the agent attempts to collect an additional mint, then the agent is killed and a loss condition is reached. Losing the game also causes the agent to lose 20 points.
A waiter (not seen in Figure \ref{fig:wfmLevel1}) moves around the board distributing mints at random. This means it is possible for an agent to get trapped while the waiter places mint on the agent's square, forcing the agent to eat it. The agent must, therefore, try to avoid getting trapped.



The optimal strategy 
is to collect as many mints as possible without collecting too many, which is currently set as nine. The player should avoid mints early on and try to avoid getting trapped. Near the end of the game, the agent should then eat the remaining mints to get to 9. 

\subsubsection{Goal}
WaferThinMints is our primary example of the changing heuristic deception. The mint goes from providing a positive reward to giving a substantial negative reward with the only visual indication being a green bar on the avatar that represents how full the character is. The agent must learn that the value of the mints is dependent on that green bar. Since the bar moves with the Avatar, it cannot just memorize a fixed state in which to stop eating the mints. The mint is distributed by a chef and left around the board at random. For the agent to play optimally, it should also learn that it is not good to get full early on because it might get trapped in and forced to eat another mint at some point.

\subsubsection{Results}
As can be seen from the graph, this agent did not have enough time to converge completely. This points to the difficulty of learning in the noisy environment where even a good strategy could result in a bad reward if the agent is unlucky. This is necessary though, as in a simpler environment with a fixed mint layout, the agent would learn to memorize a path that results in a perfect score. The agent shows some improvement over time but still plays very poorly.

By observing the agent, we see that the agent uses location to solve this problem. At the beginning of the episode, the agent rushes to the room where the initial mints are placed. This is a guaranteed source of rewards. The agent will mostly stay in the room, a safe place, unless chased out by the chef's mint placement. After the initial mints, the agent attempts to avoid mints until it's trapped by them. 

It is not clear whether the agent understands its fullness bar or uses the amount of mints placed in the game to assess the risk of eating more mints. The agent seems to have learned that the mints become dangerous, but it seems to use strange state and location information to help it know when to eat mints. This is related to the behavior we see in the game Invest. It also is incapable of reasoning about waiting until the end of the game to eat mints when it is safer to eat, an instance of the delayed gratification deception.

\subsection{Flower (Flow)}

\begin{figure}[t]
    \centering
    \includegraphics[width=.8\linewidth]{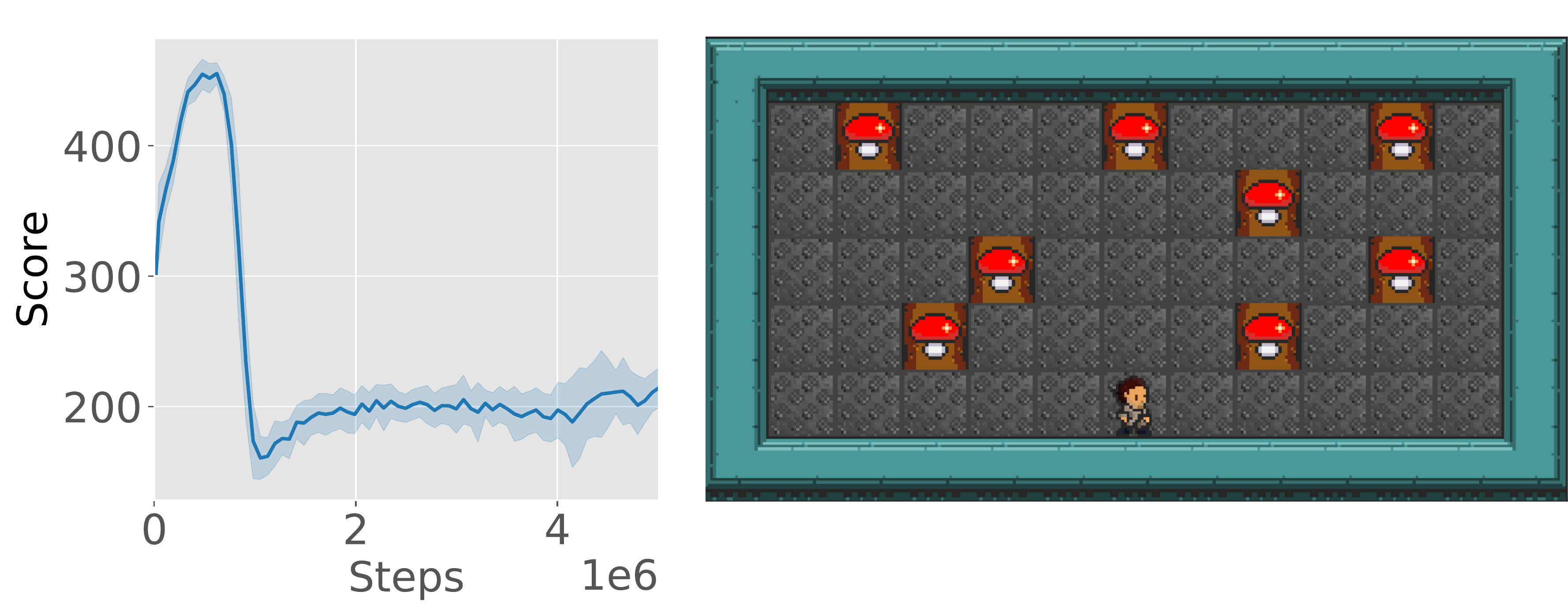}
    \caption{Flower level 1}
    \label{fig:flower}
\end{figure}

\subsubsection{Game} Flower is a game which rewards patient agents by offering the opportunity to collect a small number of points immediately, but which will grow larger over time the longer it is not collected. As shown in figure \ref{fig:flower}, a few seeds are available for the agent to collect, which are worth zero points. The seeds will eventually grow into full flowers and their point values grow along with them up to ten points. Once a flower is collected, another will begin to grow as soon as the agent leaves the space from which it was collected. 


The optimal strategy for \textit{Flower} is to let the flowers grow to their final stage of development before collecting them. 


\subsubsection{Goal}
In Flower, an agent is rewarded every time it collects a flower. To get maximum points the agent should collect each flower the moment it matures to 10 points. This will provide a better score than constantly collecting seedlings.

\subsubsection{Results}
The training graph for this game shows the agent falling for the specific deception with the sudden drop-off in performance. As the agent gets better at knowing where the flowers are, the score starts to improve. Then the agent gets too good at collecting the flowers, and they no longer have a chance to grow, lowering the score. Watching agent replays further confirms this, the agent finds a circuit through all the flowers and then gets better at quickly moving through this circuit. The agent perfectly falls for the deceit and has no way back unless it ignores the immediate rewards.

\subsection{Invest (Inv)}

\begin{figure}[t]
    \centering
    \includegraphics[width=.8\linewidth]{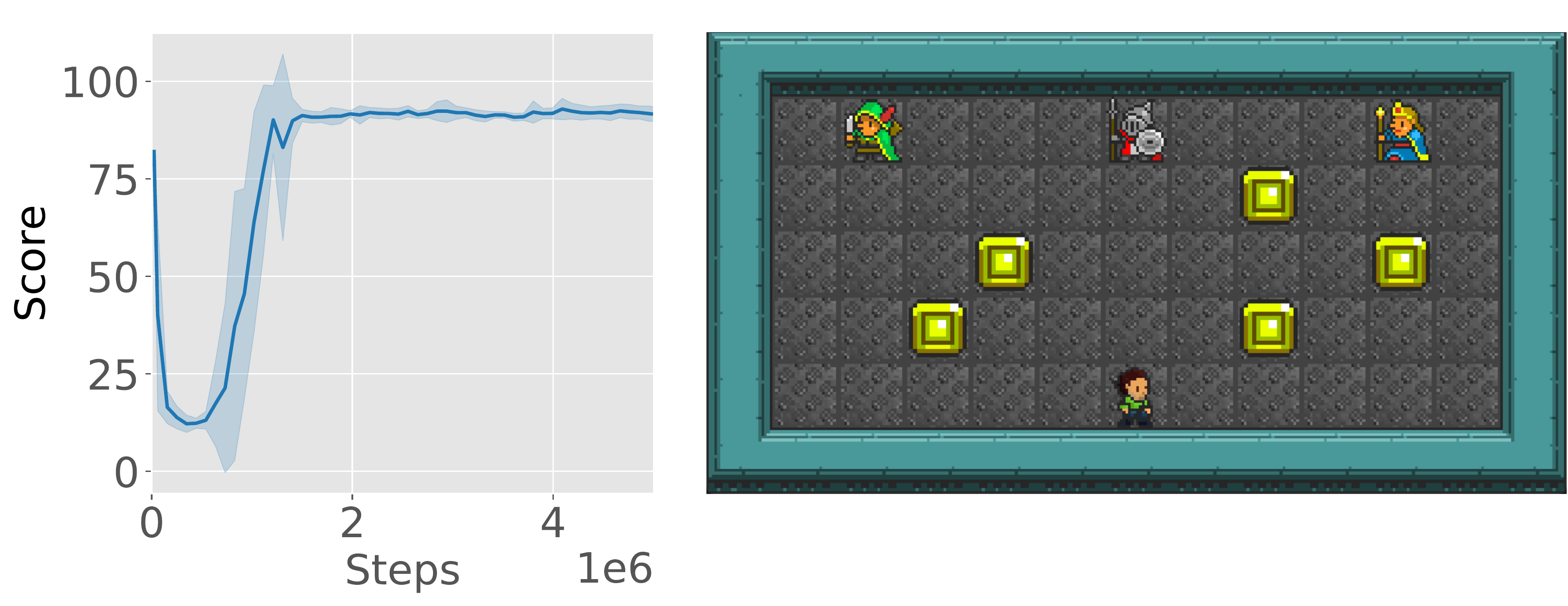}
    \caption{Invest level 1}
    \label{fig:invest}
\end{figure}

\subsubsection{Game} Invest is a game where agents can forgo a portion of their already accumulated reward, for the benefit of receiving a larger reward in the future. The level used is shown in figure \ref{fig:invest}. The agent begins with no points but can collect a small number of coins around the level to get some initial amount. These points can then be ``spent'' on certain investment options. Doing this will deduct a certain number of points from the agent's current score, acting as an immediate penalty, but will reward them with a greater number of points after some time has passed. The agent has several different options on what they can invest in, represented by the three human characters (referred to as bankers) in the top half of the level. Each banker has different rules: Green banker turns 3 into 5 after 30 ticks, Red turns 7 into 15 after 60 ticks, and Blue turns 5 into 10 after 90 ticks. The agent can decide to invest in any of these bankers by simply moving onto them, after which the chosen banker will take some of the agent's points and disappear, returning a specific number of timesteps later with the agent's reward. The agent will win the game once the time limit for the level expires. 


The optimal strategy for \textit{Invest} is defined as successfully investing with everyone as often as possible. 

\subsubsection{Goal}
Invest is a game where the agent has to intentionally seek some negative reward to get a positive reward, and then wait for a certain amount of time to get the positive reward. This delayed reward makes it very difficult for the reinforcement learning algorithm to assign credit to a specific assignment. The initial investment will only be assigned a negative reward, and the agent then has to figure out that the reward that happens later should also be assigned to this action. 
In this case, the reward is deterministic, and the challenge could be increased further by making the delay stochastic.

\subsubsection{Results}
The agent learns a very particular strategy for all five instances of training. The agent first collects all the coins and then invests with the Green Banker. From there it runs to the far right corner and waits, some agents always choose the top while others choose the bottom. As soon as the Green banker returns, the agent runs back over and reinvests only to run back to its corner and wait. This at first seems like puzzling behavior as a better strategy would be to sit next to the Green Banker and be able to reinvest faster and collect more points. On closer inspection, it becomes apparent that the time it takes the agent to reach the far corner correlates with the arrival of the delayed reward. It appears that the agent learned that investing in the Green Banker and then touching the far tile resulted in a large positive reward.

The size of the game board allowed the agent to embody the delay through movement and predict the arrival of the reward through how long it takes to walk across the board. It is possible that the agent would have learned to invest with the other bankers if the board was larger so the agent could have found a location associated with the delayed reward.

The training graph shows an interesting story too. The initial random agent would accidentally invest with all three bankers and get a fairly high score despite not consistently investing with anyone. The agent quickly learns to avoid the negative reward associated with the bankers and its score drops. It stops investing with the Blue Banker first, then the Red, and finally the Green. After it discovers how to predict the delayed reward for the Green Banker, it starts doing this more regularly until its performance converges.

\section{Comparison with planning algorithms}

\begin{table}
\centering
\resizebox{\linewidth}{!}{
\begin{tabular}{|l||cccccc|}
\hline
\textbf{Agent} & \textbf{DC 1} & \textbf{DC 2} & \textbf{DC 3} & \textbf{Inv} & \textbf{Flow} & \textbf{Mints}\\
\hline
\hline
\textbf{aStar} & \cellcolor{blue!33}3.36 & \cellcolor{blue!44}3.54 & \cellcolor{blue!16}1.33 & \cellcolor{blue!4}17.53 & \cellcolor{blue!42}604.99 & \cellcolor{blue!10}1.92\\
\textbf{greedySearch} & \cellcolor{blue!50}5.0 & \cellcolor{blue!37}3.0 & \cellcolor{blue!15}1.23 & \cellcolor{blue!0}1.0 & \cellcolor{blue!0}6.83 & \cellcolor{red!8}-5.15\\
\textbf{sampleMCTS} & \cellcolor{blue!20}2.0 & \cellcolor{blue!25}2.0 & \cellcolor{blue!25}1.99 & \cellcolor{blue!1}3.5 & \cellcolor{blue!27}392.73 & \cellcolor{blue!32}5.73\\
\hline\hline
\textbf{NovelTS} & \cellcolor{blue!21}2.1 & \cellcolor{blue!25}2.0 & \cellcolor{blue!25}2.0 & \cellcolor{blue!1}4.8 & \cellcolor{blue!21}298.51 & \cellcolor{blue!48}8.75\\
\textbf{Return42} & \cellcolor{blue!50}5.0 & \cellcolor{blue!25}2.0 & \cellcolor{blue!25}2.0 & \cellcolor{blue!43}190.12 & \cellcolor{blue!23}329.73 & \cellcolor{red!4}-2.66\\
\textbf{YBCriber} & \cellcolor{blue!50}5.0 & \cellcolor{blue!50}4.0 & \cellcolor{blue!50}4.0 & \cellcolor{blue!2}10.91 & \cellcolor{blue!21}300.73 & \cellcolor{blue!28}5.2\\
\hline\hline
\textbf{A2C} & \cellcolor{blue!50}5.0 & \cellcolor{blue!47}3.79 & \cellcolor{blue!25}2.0 & \cellcolor{blue!16}69.6 & \cellcolor{blue!16}228.86 & \cellcolor{red!10}-6.21\\
\hline
\end{tabular}
}
\caption{\small{Average score for different games using different agents. Darker blue entries have higher positive score values for that game between all the agents, while darker red entries have higher negative score values.}}
\label{tab:averageResults}
\end{table}

In this section we want to compare the results from some of the planning agents in the previous paper \cite{anderson2018deceptive} with the deep RL results in this paper. Table~\ref{tab:averageResults}, 
shows the average score respectively for all the games using six different planning agents and the trained reinforcement learning agents. Every agent plays each game around 150 times, and the average score is recorded. These are drastically different algorithms from A2C, but they provide context for how different algorithms are affected by our deceptions.

While the planning agents perform slightly better on average, this depends highly on what exact planning algorithm we are examining. The planning algorithms have an advantage over the reinforcement learning algorithm as they have a running forward model that can predict the results of each action. On the other hand, the small time frame (40 milliseconds), for deciding the next action, doesn't give the algorithm enough time to find the best action. 

In an important way, both RL and planning are facing a similar problem here. In both cases, the algorithms can only query the game environment a limited amount of times. This makes it impossible to look at all possible futures and forces the algorithms to prioritize. While most planning agents entirely rely on the given forward model, some, such Return42, also use online learning. These agents initially play with the forward model but will try to learn and generalize the game rules while playing. As the game progresses, they rely more and more on those learned abstractions. In general, this is an efficient and smart strategy but makes them vulnerable to deceptions where the game rules changed in the middle of the game, such as in \textit{Wafer Thin Mints}. Here the agents might get deceived if they do not verify the result using the forward model. This is very similar to the problem that A2C encounters since the network representation is tries to generalize the states of the game.


In summary, while the best planning agents seem to be stronger  than A2C, they also are subject to different forms of deceptions, dependent on how they are implemented.

\section{Discussion}
In summary, while the A2C deep reinforcement learning\cite{mnih2016asynchronous} approach performs somewhat well, it rarely achieves the optimal performance in our games and is vulnerable to most deceptions discussed here. In contrast, the A2C algorithm performs quite well across the board for different AI benchmarks and can be considered competitive \cite{arulkumaran2017deep,justesen2017deep}. It should also be noted that the fast-moving field of deep reinforcement learning has already produced numerous modifications that could potentially solve the games discussed here\cite{arulkumaran2017deep}. However, instead of discussing possible modifications to overcome any particular challenge presented here, we want to take a step back and refocus back on the point of this exercise. We are interested in deceptions to gain a better understanding of the general vulnerabilities of AI approaches, and try to gain a more systematic understanding of the ways deep learning in particular, and AI, in general, might fail. With the previous games as concrete examples in mind, we now want to discuss four, non-exhaustive, categories for deception. 

\subsection{Types of Deception}
\subsubsection{Lack of Hierarchical Understanding}

The DeceptiCoin games 
are relatively easy to solve if one thinks about them at the right level of abstractions. DeceptiCoins can be seen as a single binary decision between one path and another. Once this is clear, one can quickly evaluate the utility of choosing the correct one and pick the correct path. The deceptive element here is the fact that this is presented to the AI as an incredibly large search space, as it takes many steps to complete the overall meta-action. Humans are usually quite good at finding these higher levels of abstraction, and hence this problem might not look like much of a deception to us - but it is pretty hard for an AI. The large search space, paired with the assumptions that all actions along the path of the larger action matter, makes it very hard to explore all possible steps until a possible reward is reached. This is a similar problem to the famous problem in Montezuma's Revenge, where the AI could not reach the goal, and its random exploration did not even get close. This problem was only recently solved with forced exploration \cite{ecoffet2019go}. 

Finding a good hierarchical abstraction can actually solve the problem. For example, in DeceptiCoins we can look at the path from one point to another as one action - something that has been explored in GVGAI playing agents before. 

\subsubsection{Subverted Generalization}

Wafterthinmints is a game specifically designed to trick agents that generalize. Agents that simply use a forward model to plan their next step perform quite well here, as they realize that their next action will kill them. But in general, we do not have access to a forward model, so there is a need to generalize from past experience and use induction. The fact that each mint up to the 9th gives a positive rewards reinforces the idea that eating a mint will be good. The 10th mint then kills you. This is not only a problem for reinforcement learning, but has been discussed in both epistemology \cite{hume1739,Russel1912} and philosophy of AI - with the consensus that induction in general does not work, and that there is not really a way to avoid this problem. The subverted generalization is also a really good example of how more advanced AIs become more vulnerable to certain deceptions. On average, generalization is a good skill to have and can make an AI much faster, up to the point where it fails. 

\subsubsection{Delayed Reward}

The big challenge in reinforcement learning is to associate what actions lead to the reward \cite{Sutton1992}. One way to complicate this is to delay the payment of this reward, as we did in the example of invest. The player first has to incur a negative reward to invest, and then, after a certain amount of time steps gets a larger positive reward. The RL agent had two problems with Invest. First, it only ever invests with the investor with the shortest repayment time. The Red Banker would, overall, offer the best payout, but the RL agent either does not realize this relationship, or does not associate the reward correctly. 

Furthermore, the RL agents also seems to be learning ``superstitions''. When we examined the behaviour of the evolved RL agent, we see that the agent invests with the Green Banker and then runs to a specific spot in the level, waiting there for the reward payout. This behaviour is then repeated, the agent runs to the banker and then back to the spot to wait for its reward. We reran the training for the RL agent and saw the same behaviour, albeit with a different spot that the agent runs to. We assume that this superstition arose because the agent initially wandered off after investing in the Green Banker, and then received the reward when it was in that spot. It seems to have learned that it needs to invest in the banker - as varying this behaviour would result in no payout. But there is little pressure to move it away from its superstition of waiting for the result in a specific spot, even though this has no impact on the payout. In fact, it makes the behaviour, even with just the Green Banker sub-optimal, as it delays the time until it can invest again, as it has to run back to the green banker. 

What was exciting about this behavior, was the fact that similar behavior was also observed in early reinforcement learning studies with animals \cite{skinner1948superstition}. Pigeons that were regularly fed by an automatic mechanism (regardless of their behaviour) developed different superstitious behaviours, like elaborate dance and motions, which Skinner hypothesized were assumed (by the pigeon) to causally influence the food delivery. In our game, the agent seems to develop similar superstitions. 



\subsubsection{Delayed Gratification}

There is a famous experiment \cite{mischel1972cognitive} about delayed gratification that confronts 4 year old children with a marshmallow, and asks them not to eat it while the experimenter leaves the room. They are told that they will get another marshmallow, if they can just hold off eating the first marshmallow now. This task proves difficult for some children, and it is also difficult for our agent. 
Flower is a game where the agent actually gets worse over time. This is because it initially is not very good at collecting the flowers, which allows the flowers time to mature. The optimal strategy would be to wait for the flowers to grow fully, and then go around and collect them. 
The agent learns the expected reward of collecting seeds early on but does not realize that this reward changes with faster collection. When it updates its expected reward based on its new speed, it forgets that it could get higher rewards when it was slower.
While some of the planning algorithms perform better here, it is likely that they did not actually ``understand'' this problem, but are simply much worse at collecting the flowers (like the untrained RL agent). This example demonstrates that we can design a problem where the AI gets worse over time by ``learning'' to play. 


\section{Conclusion}

It appears that deep reinforcement learners are easily deceived. We have devised a set of games specifically to showcase different forms of deception, and tested one of the most widely used RL algorithms, Advantage Actor-Critic (A2C), on them. In all games, the reinforcement learners failed to find the optimal policy (with the exception that it found the optimal policy on one level of one game), as it evidently fell for the various traps laid in the levels. 

As the games were implemented in the GVGAI framework, it was also possible for us to compare with tree search-based planning agents, including those based on MCTS. (This is very much a comparison of apples and oranges, as the planning agents have access to a forward model and direct object representation but are not given any kind of training time.) We can see that for every game, there is a planning agent which performs better than the A2C agent, but that there is in most cases also a planning agent that performs worse. It is clear that some kinds of deception affect our reinforcement learning algorithm much more severely than it affects the planning algorithms; in particular, the subverted generalization of WaferThinMints. On the other hand, it performed better than most planning algorithms given the delayed reward in Invest, even though the policy it arrived at is bizarre to a human observer and suggests a warped association between cause and effect.

We look forward to testing other kinds of algorithms on these games, including phylogenetic reinforcement learning methods such as neuroevolution. We also hope that other researchers will use these games to test the susceptibility of their agents to specific deceptions.


\fontsize{9.0pt}{10.0pt} \selectfont 
\bibliography{references}
\bibliographystyle{aaai}
\end{document}